\documentclass{article}
\usepackage{spconf,amsmath,graphicx}

\title{User independent Emotion Recognition with Residual Signal-Image Network}
\name{Guanghao Yin, Shouqian Sun, Hui Zhang,  Dian Yu, Chao Li, Kejun Zhang, Ning Zou$^*$}
\address{College of Computer Science,Zhejiang University, Hangzhou, China 310027\\
\{ygh\_zju,ssq,huiz,yudian329,superli,zhangkejun,zn007\}@zju.edu.cn}

\begin{document}
%
\maketitle
\begin{abstract}
User independent emotion recognition with large scale physiological signals is a tough problem. There exist many advanced methods but they are conducted under relatively small datasets with dozens of subjects. Here, we propose Res-SIN, a novel end-to-end framework using Electrodermal Activity(EDA) signal images to classify human emotion. We first apply convex optimization-based EDA (cvxEDA) to decompose signals and mine the static and dynamic emotion changes. Then, we transform decomposed signals to images so that they can be effectively processed by CNN frameworks. The Res-SIN combines individual emotion features and external emotion benchmarks to accelerate convergence. We evaluate our approach on the PMEmo dataset, the currently largest emotional dataset containing music and EDA signals. To the best of author's knowledge, our method is the first attempt to classify large scale subject-independent emotion with 7962 pieces of EDA signals from 457 subjects. Experimental results demonstrate the reliability of our model and the binary classification accuracy of 73.65\% and 73.43\% on arousal and valence dimension can be used as a baseline$\footnote{The accepted version of The 26th IEEE International Conference on Image Processing (ICIP)}$.
\end{abstract}

\section{Introduction}
\label{sec:intro}
The emotion recognition is a research hotspot for the significant hallmark of intelligent human-computer interaction which has been extended in safe driving monitor\cite{healey1999quantifying}, mental health\cite{guo2013pervasive}, or etc.. Emotion recognition using physiological signals can guarantee the reliability because all those internal signals are reacted from the
autonomic and somatic nervous systems(ANS and SNS) which are largely involuntarily activated. Thus, signals like Electrodermal Activity(EDA) are so difficult to be controlled subjectively that they can reflect emotion objectively. Establishing the user independent relationship between emotional changes and biological signals is a challenging problem, especially in large scale dataset.

\begin{figure}
  \centering
  \includegraphics[width=79mm]{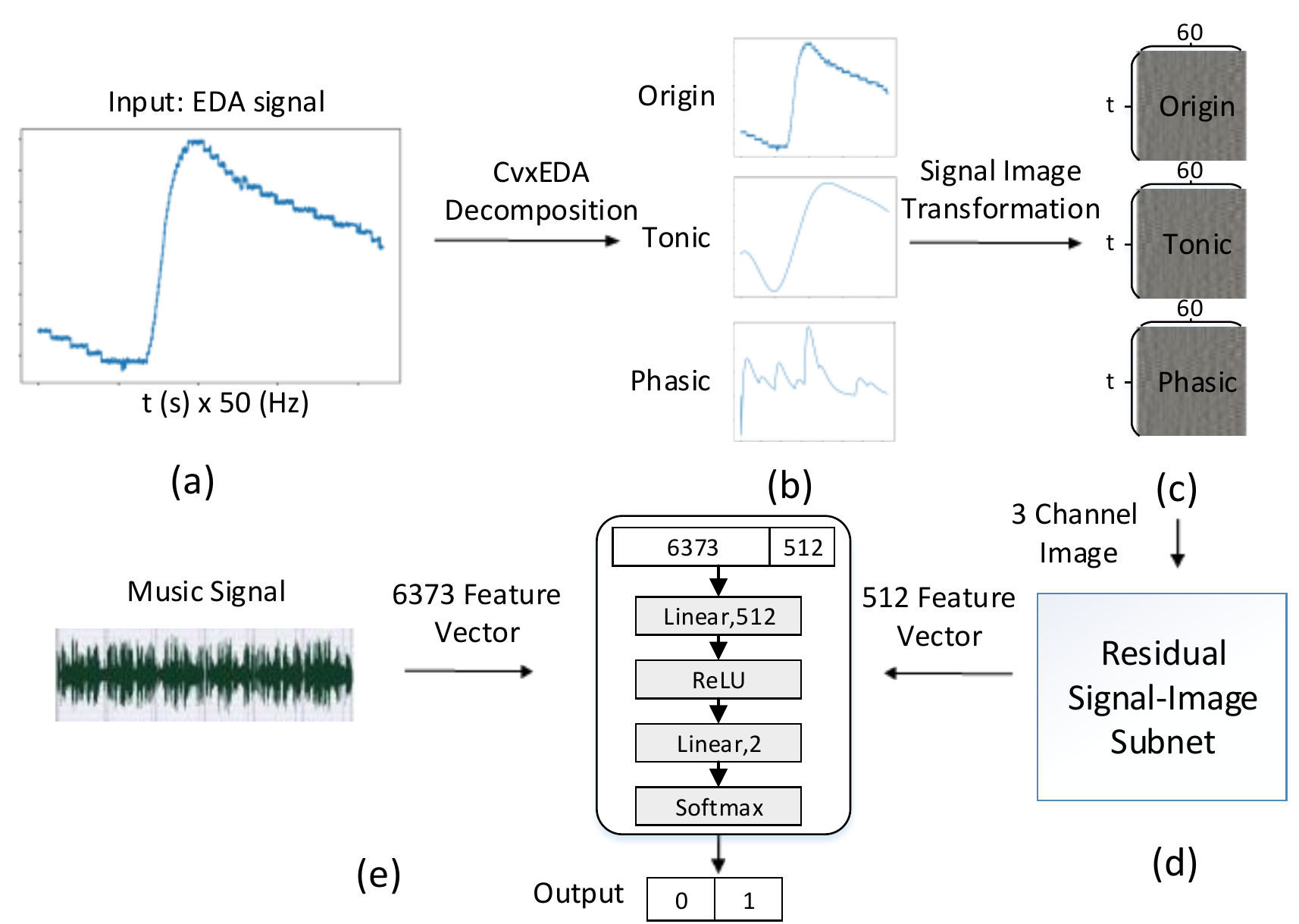}\\
  \caption{The architecture of Residual Signal-Image Network for emotion recognition and subnet$\,$(d) is shown in Fig
   \ref{fig:unit}.}
  \label{fig:model}
\end{figure}

For physiological signals based emotion recognition, two broad methods exist :$\,$the hand-crafted features
based methods with well-designed classifiers like SVM\cite{placidi2016classification}, KNN\cite{greco2017arousal}, decision tree\cite{gong2016emotion} and auto feature extraction based deep learning methods like deep belief network\cite{zheng2014eeg}, LSTM network\cite{bashivan2015learning}. All the methods have achieved good performances but they are conducted in relatively small scale datasets. Several datasets using medium stimulation have been established like DEAP\cite{koelstra2012deap}, SEED\cite{seed}, MAHNOB\cite{Soleymani2012A} etc.. 
Music and video are commonly utilized mediums of external emotion stimulation because they are effective to modulate subject's mood\cite{yang2011music,hanjalic2005affective}. In order to get good generalization performance, the model should be tested in large scale data. We test our approach with the current largest PMEmo dataset which contains 7962 EDA signals collected from 457 subjects\cite{zhang2018pmemo}.

In this work, we propose a classification strategy in PMEmo dataset and complement the baseline with our Res-SIN. Because the baseline in \cite{zhang2018pmemo} is conducted with regression methods to fit annotated V/A values, the predicted values and regression errors lack of clear practical significance. By contrast, the classification results are explicit to describe subjects' emotion states. With the strategy in \cite{yin2017cross}, we convert the emotion regression task of PMEmo dataset into binary classification problem (high/low classes) according to different values in the V/A emotional dimension\cite{posner2005circumplex}.
Inspired by the success of deep CNN in signal and image processing\cite{yu2011deep}, we transform EDA signals into images. Meanwhile, cvxEDA\cite{greco2016cvxeda} is applied for signal decomposition and data augmentation. Then processed signal images are fed into Res-SIN and external music features are added in classifier for network convergence. The results complement the baseline for emotion classification in currently largest PMEmo dataset.

\begin{figure}
  \centering
  \includegraphics[width=63mm]{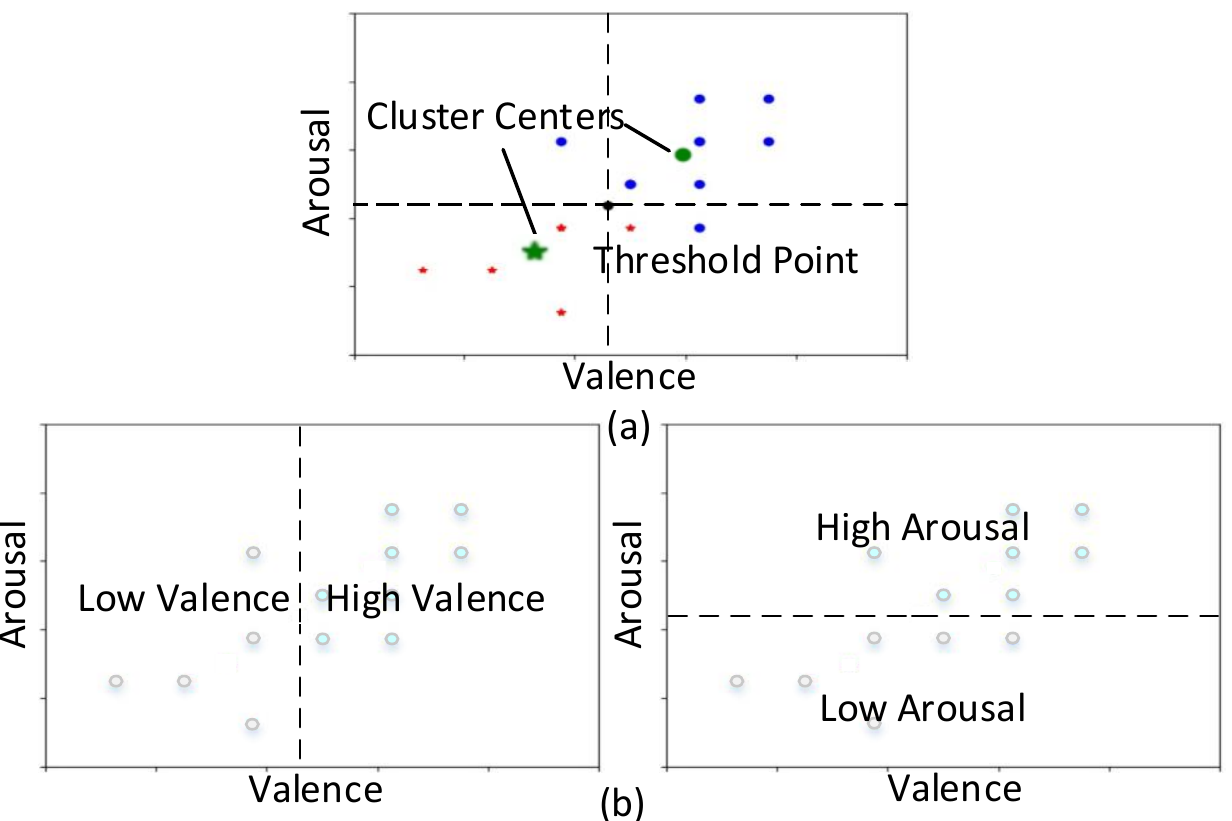}\\
  \caption{The emotion label generation of one subject:(a) The threshold point of two cluster centers based on k-means clustering; (b) The high/low valence and arousal label.}
  \label{fig:thre}
\end{figure}

\section{material and proposed method}
\label{sec:mate}
\subsection{CvxEDA Algorithm}
\label{subsec:cvx}
We focus on processing large scale EDA data. Accordingly, the strategy of EDA analysis must be efficient and convenient for processing. CvxEDA\cite{greco2016cvxeda} provides a novel method using convex optimization and prior probabilities to decompose EDA without preprocessing steps and heuristic solutions\cite{greco2017arousal}. According to \cite{boucsein2012electrodermal}, EDA can be decomposed in phasic and tonic components. Phasic signal reflects the short-term activity to stimulation. Tonic signal reflects slow drifts of the baseline skin conductance level (SCL). Taking measurement noise into account, independent and identically distributed zero-mean Gaussian noise signal is added. So the observed signal (y) is composed of three N-long column vectors: a phasic (r) and a tonic (t) signal plus the noise component ($\varepsilon$):
\begin{eqnarray}\label{equ:y}
	y = t + r + \varepsilon.
\end{eqnarray}
It can be rewrited as:
\begin{eqnarray}\label{equ:r}
  y = MA^{-1}p + B\lambda + Cd + \varepsilon.
\end{eqnarray}
where $r = MA^{-1}p$ and $t = B\lambda + Cd$. For denoising in the process of solving optimization problem, the prior probability of noise term is discarded. Finally, with the transcendental knowledge of physiology, we can get a standard Quadratic-Programming (QP) convex form(see more details in \cite{greco2016cvxeda}):
\begin{align}\label{ali:map}
	\text{min} &\frac{1}{2}\lVert MA^{-1}p + B\lambda + Cd - y \rVert ^2_{2} + \alpha \lVert MA^{-1}p \rVert _{1} + \frac{\gamma}{2} \lVert \lambda \rVert ^2_{2}\nonumber\\
	&\text{subject. to }MA^{-1}p \ge 0.
\end{align}
This QP problem can be solved efficiently with many available solvers. So signal denoising and EDA feature augmentation are accomplished at the same time.

\subsection{Classification Strategy}
In this work, we focus on classifying high/low level based in valence and arousal scales, so the label 1 indicates high V/A and the label 0 indicates low V/A. However, PMEmo dataset only provides static V/A values of different subjects' emotions which lack clear physical significance. Therefore, we convert the regression task into a binary classification and annotate the binary  emotion labels according to the continuous V/A values inspired by \cite{yin2017cross}. As we focus on user independent emotion recognition, the subject-specificity should be considered. The personal emotion threshold can be used for label generating. We calculate subjects' thresholds with the method proposed in \cite{yin2017cross} which clusters each subject's annotations by k-means clustering algorithm. Finally, labels are annotated with personal thresholds as Fig \ref{fig:thre} shows.
\begin{figure}
  \centering
  \includegraphics[width=98mm]{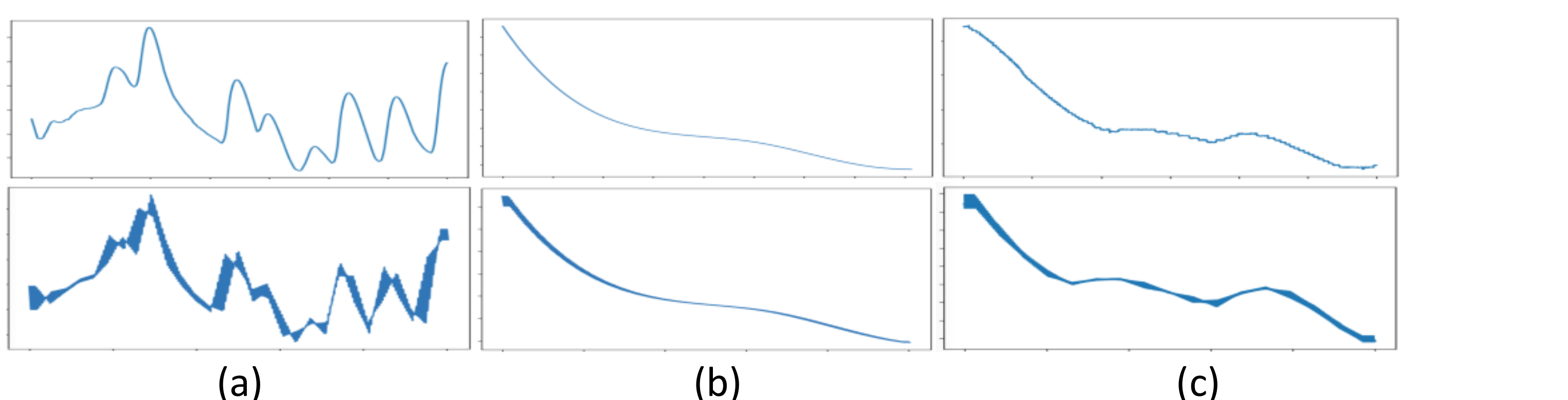}\\
  \caption{The comparison of original signals and transformed signals:$\,$(a) phasic, (b) tonic, (c) origin. First row shows original signals and next row shows transformed signals.}
  \label{fig:sig}
\end{figure}
\subsection{Proposed Res-SIN}
\label{subsec:res}
The overall architecture of Our Res-SIN (Fig \ref{fig:model}) compose of three parts as follows:

\noindent\textbf{Data Preparation }Previous research of initial orientation time\cite{aljanaki2017developing} indicates annotators need some preliminary to arousal their emotion, so we discard first 15-second signals. With CvxEDA, we extend original EDA signals to the 3 channel signals (origin, phasic and tonic) as Fig\ref{fig:model}(b) shows. Not only for denoising efficiently, the extended signal also contain more useful features. Actually, a pre-experiment was carried out and results basically validated our assumption (see more details in Section \ref{sec:exp}). After signal decomposing, we perform normalization for phasic (r), tonic (t), origin (o) signals as $A_{i,t} = (A _{i,t} - min_i)/(max_i - min_i)$ where $A_{i,t}$ is the amplitude of channel i signal at time $t$, $min_i$ and $max_i$ are separately the min/max amplitude of the channel i signal.

After data normalization, physiological signals based image tranformation is adopted. The size of original signal for one sample is $3 \times 50 \times T$, where 3 stands for the channels (r, t, o), 50 stands for the sampling rate and T stands for the duration of this signal sequence. We rearrange the original data into three gray images ($T \times 50$) and rescale them to $224 \times 224$ based on bilinear interpolation. Then random cropping and mean subtracting are conducted. When data preparation was finished, we converted signal images to 1-dimension sequences and compared them with original signals. As Fig \ref{fig:sig} shows, the signal-image transformation dosen't change the shape of curve which can guarantee the consistency of data.

\begin{figure}
  \centering
  \includegraphics[width=82mm]{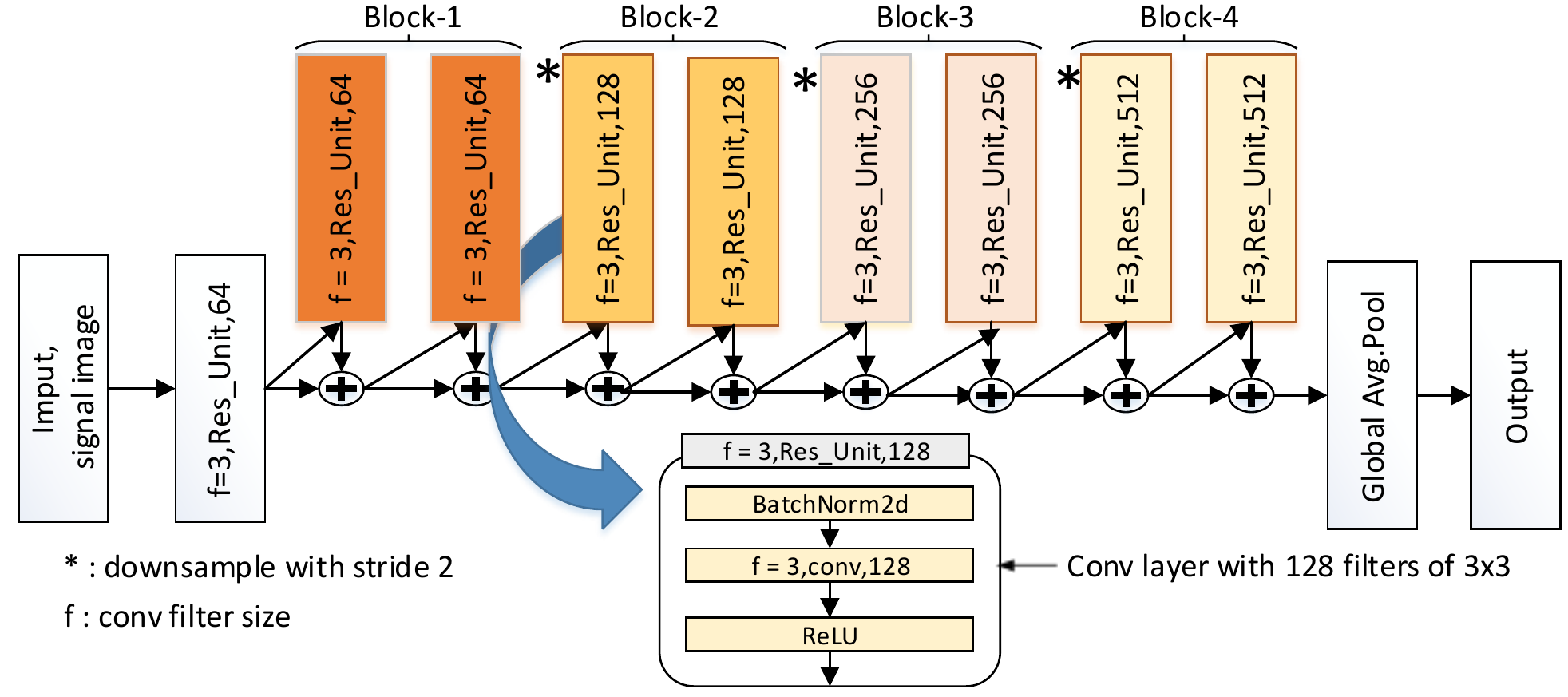}\\
  \caption{The architecture of Residual Signal-Image Subnet}
  \label{fig:unit}
\end{figure}

\noindent\textbf{Residual Signal-Image Subnet }The architecture of residual signal-image subnet is shown in Fig \ref{fig:unit}. Our signal-image subnet is built on top of Residual Neural Network \cite{he2016deep}. Considering the training efficiency and the sparsity of physiological signals, we make some changes with Resnet-18 and build our convolutional network with stacked unit of a $3 \times 3$ convolutional layer follwed by the ReLU non-linear activation function. Then, the full connected layer is discarded because this network aims for EDA feature extraction. Finally, the whole subnet outputs a 512-dimension vector.

\noindent\textbf{Multi-feature Fusion }Music is an excellent medium to raise emotion states with affection involvement, but subject-specificity determines the intricate difference of nonemotional individual contexts between the subjects\cite{kim2008emotion}. Naturally, our idea is combining general emotion benchmark from music with features of individual specificity from EDA, so we construct multilayer classifier. EDA features of the 512-dimension vector is extracted by Res-SI subnet and music feature extraction is accomplished in PMEmo dataset with a open source toolkit openSMIlE to obtain the 6373-dimension features. After z-score normalization, the music and EDA features are connected as one vector input into the multilayer classifier. A linear layer of 512-dimension output, the ReLU activation function and a linear layer of 2-dimension output are in series with the softmax layer to compose our classifier, as Fig {\ref{fig:model}} (e) shows.

We train our Res-SIN with Cross-entropy loss and empirically selected the size of mini-batches for the SGD as 100. Moreover, we set the initial learning rate to 0.001, which is decreased by multiplying it by 0.1 at every 300th iteration.
\section{Experiments}
\label{sec:exp}
In this section, three experiments will be conducted. The results and analysis can validate our idea and explain why our method achieves remarkable performance in large scale data. First, we introduce PMEmo dataset and evaluation criteria.
\subsection{Dataset And Evaluation Criteria}
\noindent\textbf{PMEmo Datase }PMEmo is the currently largest dataset\cite{zhang2018pmemo} with EDA and music signals for emotion recognition. This dataset extracted chorus parts from 794 songs as emotion stimulus, each of which was labelled by at least 10 subjects. In total, 7962 pieces of EDA signals from 457 subjects were collected at 50-Hz sampling rate. Meanwhile, subjects also annotated dynamic V/A values at a sampling rate of 2 Hz as well as a static V/A value for each music clip.In this work, we will
utilize static annotations to represent subjects’ emotion.

\noindent\textbf{Evaluation Criteria }As we convert emotion recognition to a binary classification, average classification accuracy, F1-score, precision and recall are adopted as the classifying evaluation criteria. In the pre-experiment of correlation analysing, Root Mean Square Error (RMSE) and Pearson Correlation Coefficient (r) are adopted according to the baseline in \cite{zhang2018pmemo}.

\subsection{Experiment Analysis}
\noindent\textbf{Correlation Analysis }In the original study, we assumed cvxEDA decomposition might be beneficial for useful data expansion. Hence, a pre-experiment of regression was essential to validate whether decomposed components were correlate with emotion. Referring to the methodology of \cite{zhang2018pmemo}, we also adopted Multivariate Linear Regression (MLR) and Support Vector Regression (SVR) to fit V/A static values with signals. Low-pass filtering of 0.6 HZ and z-score normalization were applied for pre-processing.

As Table \ref{cov} shows, we can make some analyses:$\,$(1) RMSE of phasic and tonic signals in valence and arousal scales are lower than the baseline which means the phasic and tonic components are both useful in emotion regression task.$\,$(2) With the lower fitting error,  phasic and tonic signals significantly improve correlation coefficient which demonstrates they have better linear correlation with emotion annotations.

These results can basically validate our assumption, moreover, motivates us to apply cvxEDA decomposition and design multi-linear classifier.

\begin{table}
\begin{center}
\caption{Regression evaluation with the baseline in PMEmo. }\label{cov}
\begin{tabular}{ccccc}
  \hline
  Method&V/A&Input&RMSE&r\\
  \hline
  MLR&Arousal&Zhang et al.\cite{zhang2018pmemo}&0.1860&0.0110\\
  &&\textbf{tonic}&\textbf{0.1760}&\textbf{0.0462} \\
  &&\textbf{phasic}&\textbf{0.1759}&\textbf{0.0395} \\
  \cline{3-5}
  &Valence&Zhang et al.\cite{zhang2018pmemo}&0.1390&0.0630 \\
  &&\textbf{tonic}&\textbf{0.1343}&\textbf{0.0699} \\
  &&\textbf{phasic}&\textbf{0.1321}&\textbf{0.0645} \\
  \hline
  SVR&Arousal&Zhang et al.\cite{zhang2018pmemo}&0.1940&0.0400 \\
  &&\textbf{tonic}&\textbf{0.1769}&\textbf{0.0537} \\
  &&\textbf{phasic}&\textbf{0.1769}&\textbf{0.0635} \\
  \cline{3-5}
  &Valence&Zhang et al.\cite{zhang2018pmemo}&0.1410&0.0170 \\
  &&\textbf{tonic}&\textbf{0.1401}&\textbf{0.0553} \\
  &&\textbf{phasic}&\textbf{0.1400}&\textbf{0.0880} \\
  \hline
\end{tabular}
\end{center}
\end{table}
\begin{figure}[htb]
  \centering
  \includegraphics[width=84mm]{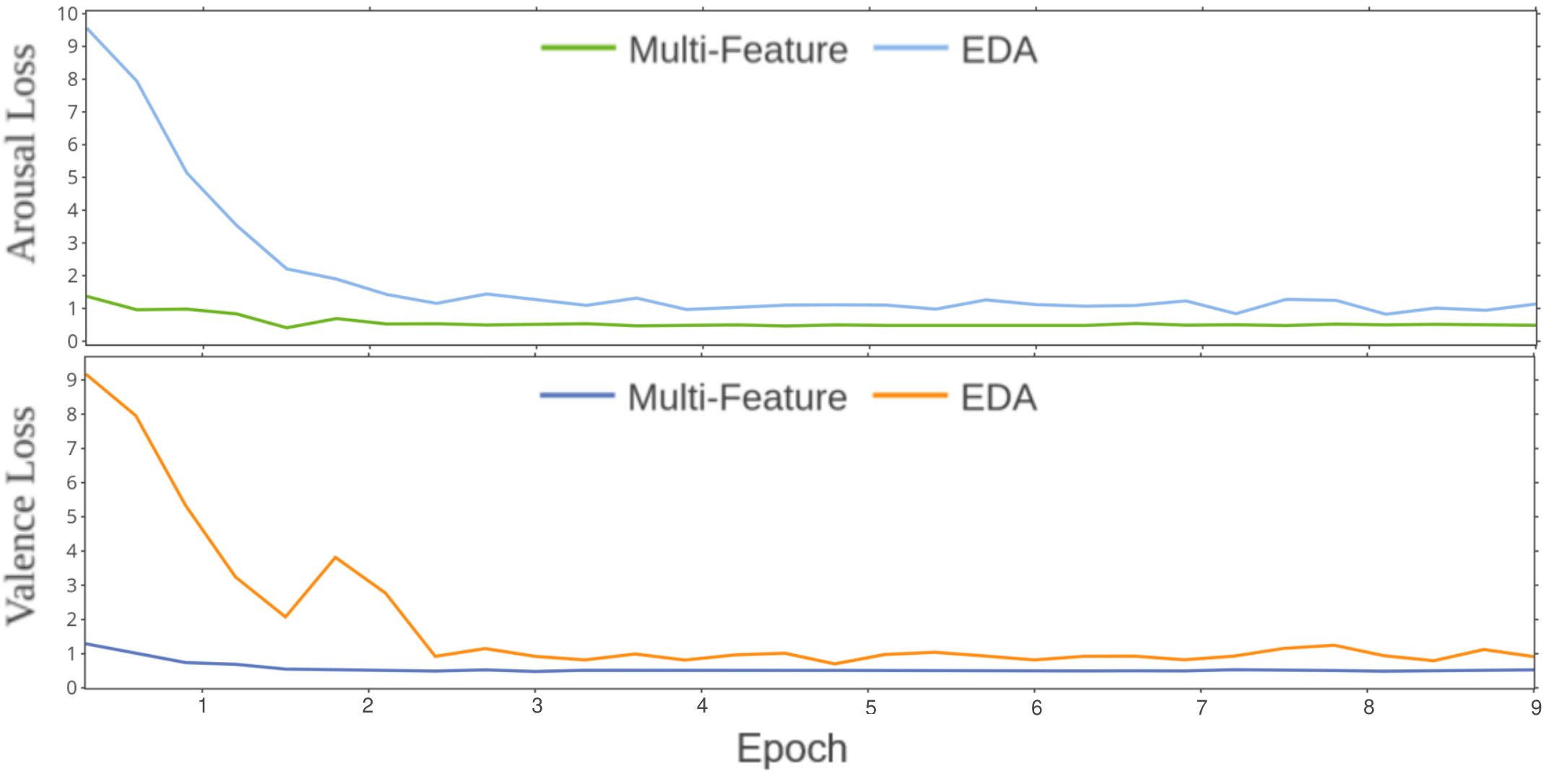}\\
  \caption{The training loss in arousal and valence scales with EDA feature and fusion feature, respectively. }
  \label{fig:loss}
\end{figure}

\noindent\textbf{Image Components Analysis }With model establishment, the experiments of high/low emotion classification in V/A dimensions should be conducted. We applied 10-fold cross-validation method. After tests of 10 flods were finished, the evaluation indexes were averaged at the end. For analysing the role every signal plays, we respectively input the phasic, tonic, 3 channel mixed signals into the model and compared results. As results shown in Table \ref{com}, we can also make some analyses: (1) Res-SIN with original signals gets the lowest accuracy due to noise interference in raw signals. (2) Res-SIN with tonic signals is steadiest and gets good F1-Score between 3 separate inputs. The tonic reflects slow drifts of SCL so it provides the stablest message of EDA trends. (3) The phasic contains the dynamic activity to provide more short-term specific messages that leads to the good accuracy. But the phasic is more intricate that gets less steady performance with lower F1-Score. (4) The mixed signals lead to the greatest performance (the highest accuracy and F1-Score).

According to the analyses above, 3 channel mixed signals get the best reliability for the fusion of dynamic and static emotion features.

\noindent\textbf{Multi-feature Fusion Analysis }As section \ref{subsec:res} explained, Res-SIN combines the subject-specificity emotion features from EDA and the emotion benchmarks from music. In order to demonstrate the efficiency of our multi-feature fusion method, we compared it with the method which separately conducted with EDA features and music features. We fed EDA images into our Res-SI subnet and music features into the SVM classifier to compared their classifying accuracies with the multi-feature baseline.

As Table \ref{com} shows, the multi-feature fusion method achieved best performance and the benchmark extracted from music adds constraints to improve recognition rate. However, it should be emphasized: the accuracy of SVM + Music method highly depents on subjects' annotation. For different people, the same music might lead to completely opposite emotion state which means the same music could map to label 0 and 1. Therefore, music features are actually meaningless out of subjective specificity in user-independent emotion classification. As Fig \ref{fig:loss} shows, the training loss of Res-SIN with fusion features decreases lower and faster than adoptting with EDA features. Because the added emotion benchmark from music feature provides more constraints, fusion feature can effectively accelerate convergence and decrease training loss.

Therefore, the great performance of our model attributes to:$\,$(1) the CvxEDA decomposition method to augment the EDA data; (2) the fusion of dynamic and static emotion activity from EDA; (3) the fusion of emotion benchmark from music and individual specificity feature from EDA.

\begin{table}
\begin{center}
\caption{Experiment results:(1) using different signal components; (2) using different features.}\label{com}
\begin{tabular}{cccccc}
  \hline
  V/A&Input&Accuracy&F1-Score&Precision&Recall\\
  \hline
  V&origin&66.19\%&74.59\%&86.02\%&65.83\% \\
  &tonic&66.76\%&75.20\%&87.35\%&66.01\% \\
  &phasic&68.41\%&72.61\%&72.59\%&72.64\% \\
  &\textbf{mix}&\textbf{73.43\%}&\textbf{77.54\%}&\textbf{83.64\%}&\textbf{72.27\%} \\
  \hline
  A&origin&70.80\%&77.34\%&84.78\%&71.11\% \\
  &tonic&71.24\%&78.22\%&87.88\%&70.48\% \\
  &phasic&73.26\%&77.97\%&80.51\%&75.59\% \\
  &\textbf{mix}&\textbf{73.65\%}&\textbf{78.56\%}&\textbf{82.12\%}&\textbf{75.29\%} \\
  \hline
\end{tabular}
\end{center}
\begin{tabular}{cccc}
  \hline
  V/A&Method and Feature&Accuracy&F1-Score\\
  \hline
  V&Res-SI Subnet + EDA &55.92\%&58.83\%\\
  &SVM + Music&70.43\%&75.32\%\\
  &\textbf{Res-SIN + Multi-feature}&\textbf{73.43\%}&\textbf{77.54\%}\\
  \hline
  A&Res-SI Subnet + EDA&57.24\%&60.12\%\\
  &SVM + Music&71.49\%&76.36\%\\
  &\textbf{Res-SIN + Multi-feature}&\textbf{73.65\%}&\textbf{78.56\%}\\
  \hline
\end{tabular}
\end{table}

\section{conclusion}

In this paper, we propose an end-to-end CNN framework (Res-SIN) for user-independent emotion recognition. For the specific significance of research, we convert emotion recognition to a binary classification. For conducting deep CNN, the 1-dimension EDA signals are transformed to 2-dimension images. Meanwhile, a novel data augmentation method of cvxEDA is conducted. The fusion of subject specificity features from EDA and benchmarks from audio further improves the accuracy of recognition which demonstrates the effectiveness of our network structure and the complementarity of multiple features. Experiment results achieve great performance and complements the baseline of emotion classification in PMEmo dataset which also demonstrates the fusion of general benchmark and subject specificity is an efficient way in future large-scale emotion recognition.

\bibliographystyle{IEEEbib}
\bibliography{strings,refs}

\end{document}